\documentclass[10pt,twocolumn,letterpaper]{article}

\usepackage[pagenumbers]{cvpr} 

\usepackage{marvosym}  
\usepackage{fancyhdr}

\definecolor{cvprblue}{rgb}{0.21,0.49,0.74}
\usepackage[pagebackref,breaklinks,colorlinks,allcolors=cvprblue]{hyperref}
\usepackage{multirow}
\usepackage{adjustbox}
\usepackage[table]{xcolor}
\usepackage{amssymb}

\newlength{\yuvionheadextra}
\fancypagestyle{yuvionheader}{%
  \fancyhf{}%
  \fancyhead[L]{\raisebox{0pt}{\includegraphics[height=17pt]{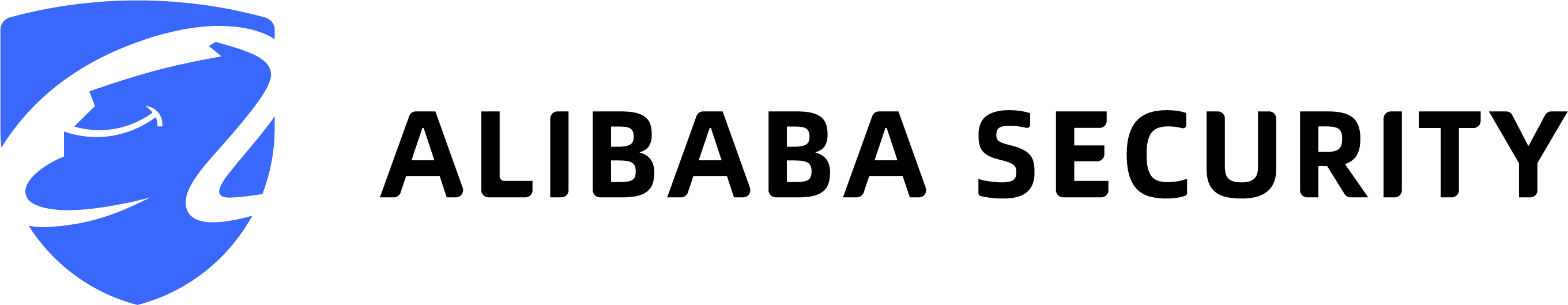}}}%
  \fancyhead[C]{}%
  \fancyhead[R]{\raisebox{0pt}{\includegraphics[height=17pt]{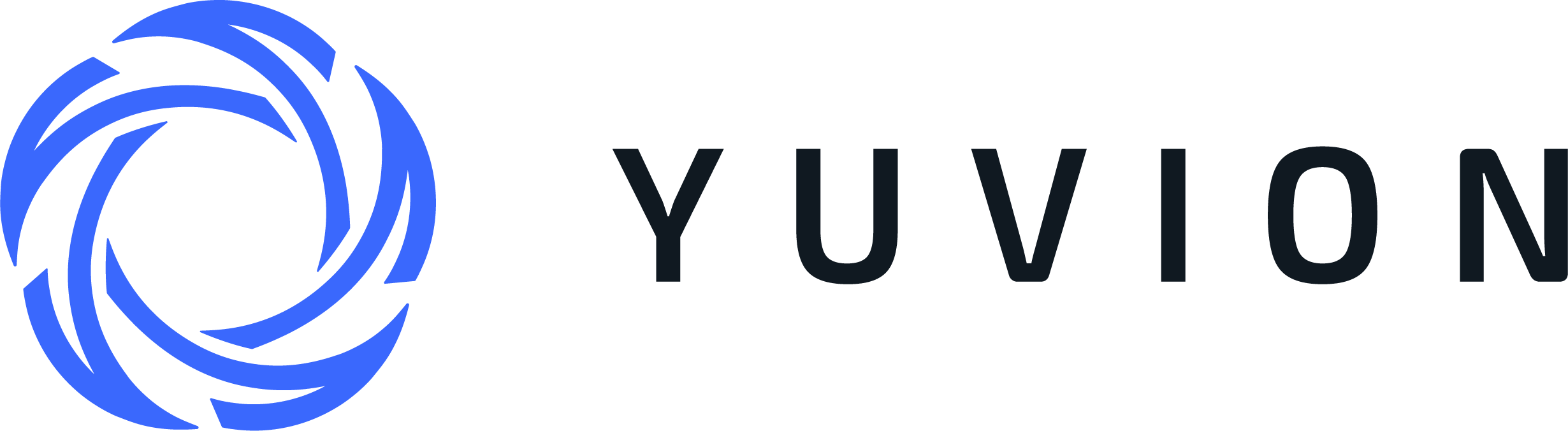}}}%
  \fancyfoot[C]{\thepage}%
}

\def\confName{CVPR}
\def\confYear{2026}

\title{TC-Padé: Trajectory-Consistent Padé Approximation for Diffusion Acceleration}

\author{%
Benlei Cui$^{1\text{*}}$ \quad
Shaoxuan He$^{2\text{*}}$ \quad
Bukun Huang$^{3}$ \quad
Zhizeng Ye$^{3}$ \quad
Yunyun Sun$^{1}$ \quad
Longtao Huang$^{1}$ \\
Hui Xue$^{1}$ \quad
Yang Yang$^{1}$ \quad
Jingqun Tang$^{4}$ \quad
Zhou Zhao$^{2,\text{\Letter}}$ \quad
Haiwen Hong$^{1,\text{\Letter}, \dagger}$
\\[0.8em]
$^{1}$Yuvion Team, Alibaba Group \quad
$^{2}$Zhejiang University \quad \\
$^{3}$Fireflyfusion, QuVideo Inc \quad
$^{4}$ByteDance Intelligent Creation
\\[0.2em]
{\small \texttt{\{cuibenlei.cbl, heshaoxuan.hsx, honghaiwen.hhw\}@alibaba-inc.com}} \\
}

\begin{document}

\maketitle
\pagestyle{yuvionheader}
\thispagestyle{yuvionheader}

\begingroup
\renewcommand{\thefootnote}{\fnsymbol{footnote}}
\footnotetext{%
  $^{*}$Equal contribution. \quad
  $^{\text{\Letter}}$Corresponding authors. \quad
  $^{\dagger}$Project lead.
}
\endgroup

\begin{abstract}

Despite achieving state-of-the-art generation quality, diffusion models are hindered by the substantial computational burden of their iterative sampling process. While feature caching techniques achieve effective acceleration at higher step counts (e.g., 50 steps), they exhibit critical limitations in the practical low-step regime of 20-30 steps. As the interval between steps increases, polynomial-based extrapolators like TaylorSeer suffer from error accumulation and trajectory drift. Meanwhile, conventional caching strategies often overlook the distinct dynamical properties of different denoising phases.   To address these challenges, we propose Trajectory-Consistent Padé(\textbf{TC-Padé}) approximation, a feature prediction framework grounded in Padé approximation. By modeling feature evolution through rational functions, our approach captures asymptotic and transitional behaviors more accurately than Taylor-based methods. To enable stable and trajectory-consistent sampling under reduced step counts, TC-Padé incorporates (1) adaptive coefficient modulation that leverages historical cached residuals to detect subtle trajectory transitions, and (2) step-aware prediction strategies tailored to the distinct dynamics of early, mid, and late sampling stages. Extensive experiments on DiT-XL/2, FLUX.1-dev, and Wan2.1 across both image and video generation demonstrate the effectiveness of TC-Padé. For instance, TC-Padé achieves 2.88$\times$ acceleration on FLUX.1-dev and 1.72$\times$ on Wan2.1 while maintaining high quality across FID, CLIP, Aesthetic, and VBench-2.0 metrics, substantially outperforming existing feature caching methods.

\end{abstract}

\section{Introduction}
\label{sec:intro}
Diffusion models~\cite{ho2020denoising,sohl2015deep,song2020score} have emerged as the dominant paradigm in generative artificial intelligence, achieving state-of-the-art performance across diverse tasks including image synthesis~\cite{flux2024,esser2024scaling} and video generation~\cite{kong2024hunyuanvideo,wan2025wan}. The introduction of Diffusion Transformers (DiT)~\cite{peebles2022scalable} has substantially enhanced model expressiveness and generation quality through increased scale and capacity. However, the iterative nature of denoising process requires over tens to hundreds of sequential network evaluations. This computational burden poses a critical bottleneck for practical deployment in latency-sensitive and resource-constrained scenarios.

To mitigate these costs, a range of acceleration techniques has emerged. Among these approaches, caching-based strategies have gained traction for their training-free, plug-and-play nature~\cite{liu2025survey}. Early caching methods such as DeepCache~\cite{ma2024deepcache} and Block Caching~\cite{wimbauer2024cache} leverage architectural skip connections in U-Net-based diffusion models~\cite{ronneberger2015u} to reuse intermediate activations. Recent efforts to accelerate transformer-based architectures can be broadly classified into two categories. The first category is reuse-based methods, approaches like FORA~\cite{selvaraju2024fora} and $\Delta$-DiT~\cite{chen2024delta} directly adapt cache-and-reuse paradigms to DiTs, while token-level methods such as ToCa~\cite{zou2024accelerating} achieve finer-grained selective reuse by decomposing features and caching only informative components. The second category is prediction-based methods, which have recently emerged as the state-of-the-art paradigm by shifting from passive reuse to active feature extrapolation. Most notably, TaylorSeer~\cite{liu2025reusing} introduces a polynomial-based framework that employs truncated Taylor expansion to explicitly extrapolate features to future timesteps.

\begin{figure*}[t]
\centering
\includegraphics[width=1.0\textwidth]{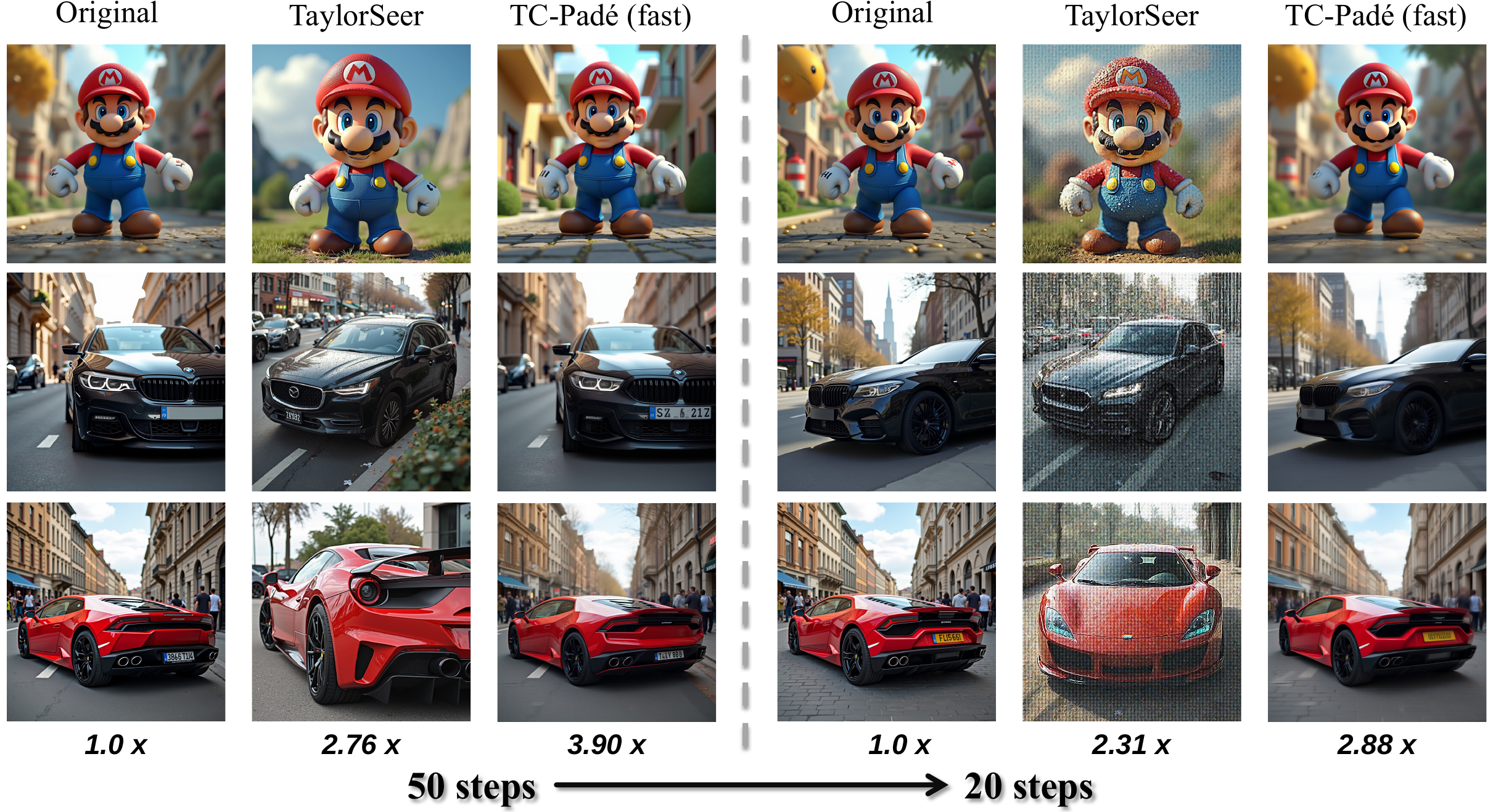}
\caption{
\textbf{Visual quality comparison on FLUX.1-dev with 50 steps (left) and 20 steps (right).} Numbers indicate speedup ratios relative to the original sampling. Previous cache-based methods exhibit notable quality degradation with altered textures and colors under low-step settings. In contrast, TC-Padé preserves visual fidelity while achieving higher acceleration. Additional qualitative comparisons with other methods are provided in the Appendix.
}
\label{fig:trajectory_consistency}
\end{figure*}

\begin{figure}
\centering
\includegraphics[width=0.5\textwidth]{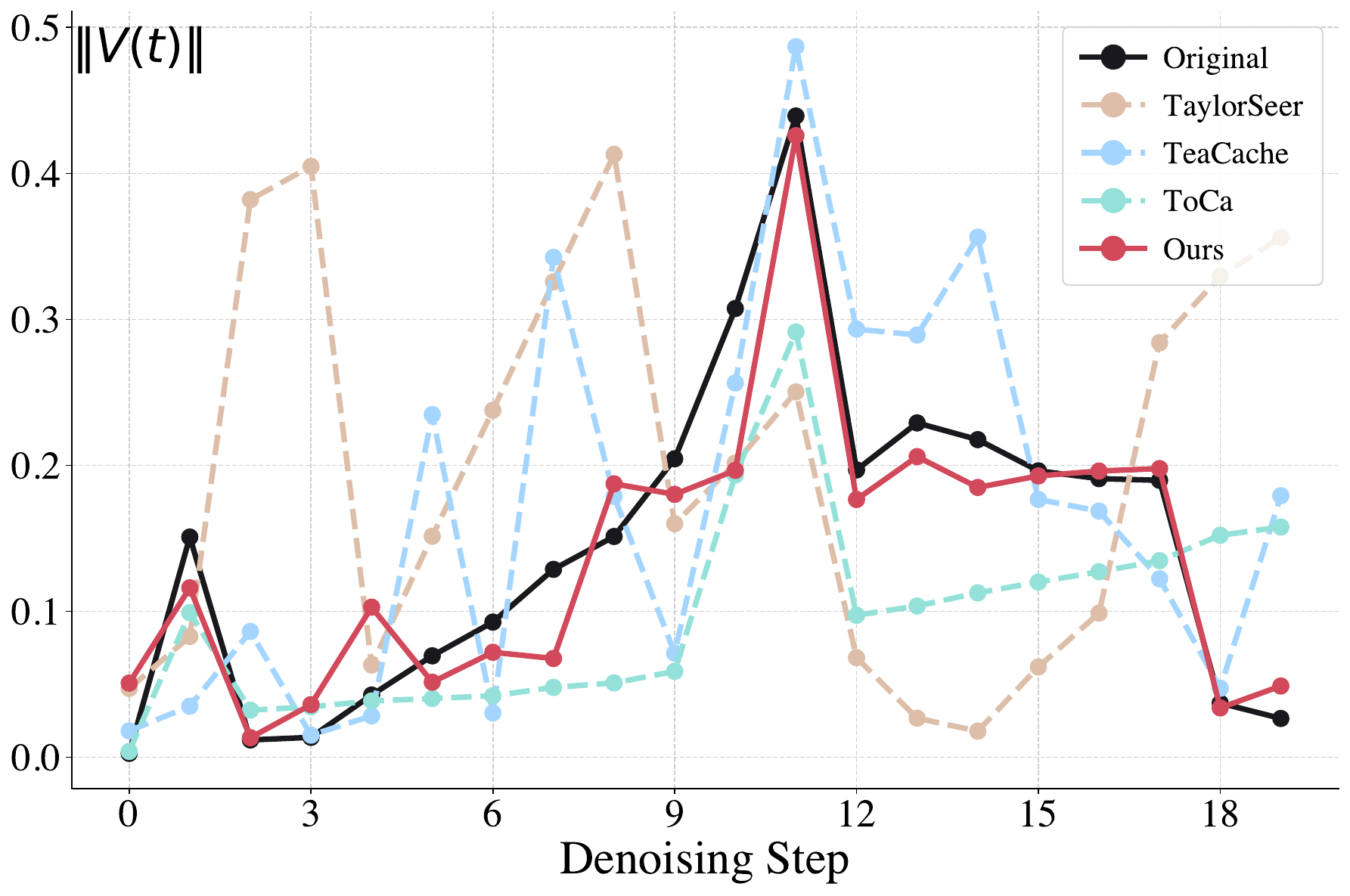}
\caption{
PCA visualization of final layer outputs from various caching-based methods under 20 steps sampling regime. $\|{V(t)}\|$ represents the PCA results of the model's output velocity field. Data is collected from FLUX.1-dev on the DrawBench~\cite{saharia2022photorealistic} dataset.
}
\label{fig:pca_visualization}
\end{figure}

Despite their promising speedups, existing feature caching methods exhibit a fundamental limitation: their efficacy degrades substantially in the 20-30 step range—a commonly adopted sampling budget in industry-grade applications~\cite{li2025ragdiffusion,gao2025postermaker}. As shown in Figure~\ref{fig:trajectory_consistency}, when employing a reduced number of denoising steps, current approaches suffer significant visual quality degradation. \textit{When the total number of denoising steps decreases, the temporal interval between consecutive evaluations increases correspondingly, causing feature similarity across timesteps to decay exponentially.} For reuse-based methods, this widened gap violates the core assumption of near-identical features~\cite{liu2025timestep,ma2025magcache}, making cached activations poorly aligned with the current state and causing severe trajectory drift. For prediction-based approaches relying on polynomial extrapolation, the enlarged cache interval amplifies approximation error due to the limited radius of convergence inherent to Taylor series. As evidenced by the PCA visualization in Figure~\ref{fig:pca_visualization}, these caching methods fail to preserve trajectory consistency across the sampling process, exhibiting substantial deviations from the ground truth trajectory. Furthermore, current approaches use the same prediction strategy throughout the entire denoising process, overlooking how features evolve differently across sampling phases. This uniform treatment becomes problematic at large timestep intervals, where each phase has distinct characteristics that need differentiated strategies.

In this paper, we propose \textbf{TC-Padé}: a trajectory-consistent residual prediction framework grounded in Padé approximation, which accelerates diffusion sampling while preserving quality under low-step regimes with large timestep intervals. Unlike Taylor series, which are constrained by finite radius of convergence and exhibit divergent behavior beyond local neighborhoods, Padé approximants—expressed as ratios of polynomials—provide superior asymptotic properties and can faithfully capture both smooth transitions and abrupt regime changes in feature dynamics~\cite{brezinski1996extrapolation}. This mathematical advantage is particularly critical for diffusion models operating with efficient sampling configurations (i.e., reduced sampling steps necessitating larger timestep intervals), where feature evolution exhibits highly nonlinear, phase-dependent behavior that polynomial extrapolation fundamentally cannot represent.

Extensive experiments on text-to-image, text-to-video and class-conditional image generation demonstrate the effectiveness of TC-Padé on FLUX.1-dev~\cite{flux2024}, Wan2.1~\cite{wan2025wan} and DiT-XL/2~\cite{peebles2022scalable} over previous feature caching methods. At 20 denoising steps, we achieve up to 2.88$\times$ speedup on FLUX.1-dev, 1.72$\times$ on Wan2.1 and 1.46$\times$ on DiT-XL/2, while maintaining high quality with only a 3\% loss in FID on FLUX.1-dev and a 4\% loss in VBench-2.0 score~\cite{zheng2025vbench}, significantly outperforming existing feature caching approaches.

In summary, our contributions are as follows:

\begin{itemize}
\item Padé-Inspired feature forecasting. We introduce TC-Padé, a rational predictor grounded in Padé approximation that faithfully models asymptotic behavior and phase transitions in feature dynamics, thereby enabling trajectory-consistent sampling even in challenging reduced-step scenarios.


\item Denoising step-aware prediction. We design timestep-aware strategies tailored to the early, middle, and late sampling stages. This innovation allows for robust prediction under large timestep intervals.


\item Comprehensive experiment validation. Extensive experiments on class-conditional image generation, text-guided image and video generation demonstrate that our method achieves substantial efficiency gains while maintaining consistent generation quality compared to the existing feature caching methods.


\end{itemize}
\section{Related Work}
\subsection{Visual Generation Acceleration}
Visual generation is currently dominated by autoregressive~\cite{chen2018pixelsnail,sun2024autoregressive} and diffusion~\cite{ho2020denoising,rombach2022high}. Beyond generic synthesis, the rapid development of diffusion models has enabled a broad range of applications, including object removal~\cite{sun2025attentiveeraser,liu2025erasediffusion}, early prediction of generation quality~\cite{cui2026diffusionprobe}, product-poster generation~\cite{cui2026simpleposter}, and virtual footwear try-on~\cite{li2025shoefit}. Autoregressive models are constrained by the computational cost of long visual token sequences~\cite{he2024zipar,he2025neighboring}, whereas diffusion models achieve remarkable generation quality at the cost of excessively long denoising steps. Prior work address this inefficiency from two complementary directions: reducing the number of sampling steps and decreasing the per-step computational cost. In the first direction, solver-based methods formulate the reverse diffusion process as the numerical integration of reverse-time ODEs or SDEs, enabling advanced numerical solvers with fewer function evaluations, such as DDIM~\cite{song2020denoising}, DPM-Solver~\cite{lu2022dpm}. To further reduce sampling steps to fewer than 10 steps, distillation-based methods train student models to predict multiple teacher steps in one pass, including progressive distillation~\cite{salimans2022progressive}, consistency models~\cite{song2023consistency}, and adversarial distillation methods~\cite{sauer2024adversarial, sauer2024fast}. For the reducing per-step computation direction, model compression techniques employ pruning~\cite{wang2023patch, zhu2024dip}, quantization~\cite{kim2025ditto, li2023q, shang2023post}, and knowledge distillation~\cite{bandyopadhyay2025sd3} to reduce the denoising network's computational footprint. Dynamic inference strategies include early exiting~\cite{tang2023deediff, moon2024simple}, feature caching~\cite{ma2024deepcache}, and token reduction~\cite{pmlr-v267-lu25v, saghatchian2025cached} that exploit redundancy across denoising steps.

\subsection{Diffusion Cache Mechanism}
Caching mechanisms emerge as a promising direction for diffusion acceleration. Unlike previous methods that require training or modifying the model archetecture, cached-based methods leverage the high temporal similarity of features between adjacent denoising steps. Early caching methods like Faster Diffusion~\cite{li2024faster} and Deepcache~\cite{ma2024deepcache} focus on caching the features by outputs of specific U-Net blocks~\cite{ronneberger2015u}. As for DiT models, techniques such as FoRA~\citep{selvaraju2024fora} and PAB~\citep{zhao2024real} cache attention features for future reuse. While $\Delta$-DiT~\citep{chen2024delta} and BlockDance~\citep{zhang2025blockdance} focus on reusing block features to skip the computation of certain blocks. ToCa~\citep{zou2024accelerating} and Tokencache~\citep{lou2024token} achieves effective acceleration by innovatively shifting the caching target to tokens, thereby reducing information loss. TeaCache~\citep{liu2025timestep} predicts output change and gate reuse, by utilizing the differences in the noise input through time-step embedding. Recent work introduces TaylorSeers~\citep{liu2025reusing}, which leverages a Taylor-series expansion of the diffusion dynamics to extrapolate feature representations at the subsequent timestep.

\section{Method}
\label{sec:method}

\begin{figure*}[t]
\centering
\includegraphics[width=1.0\textwidth]{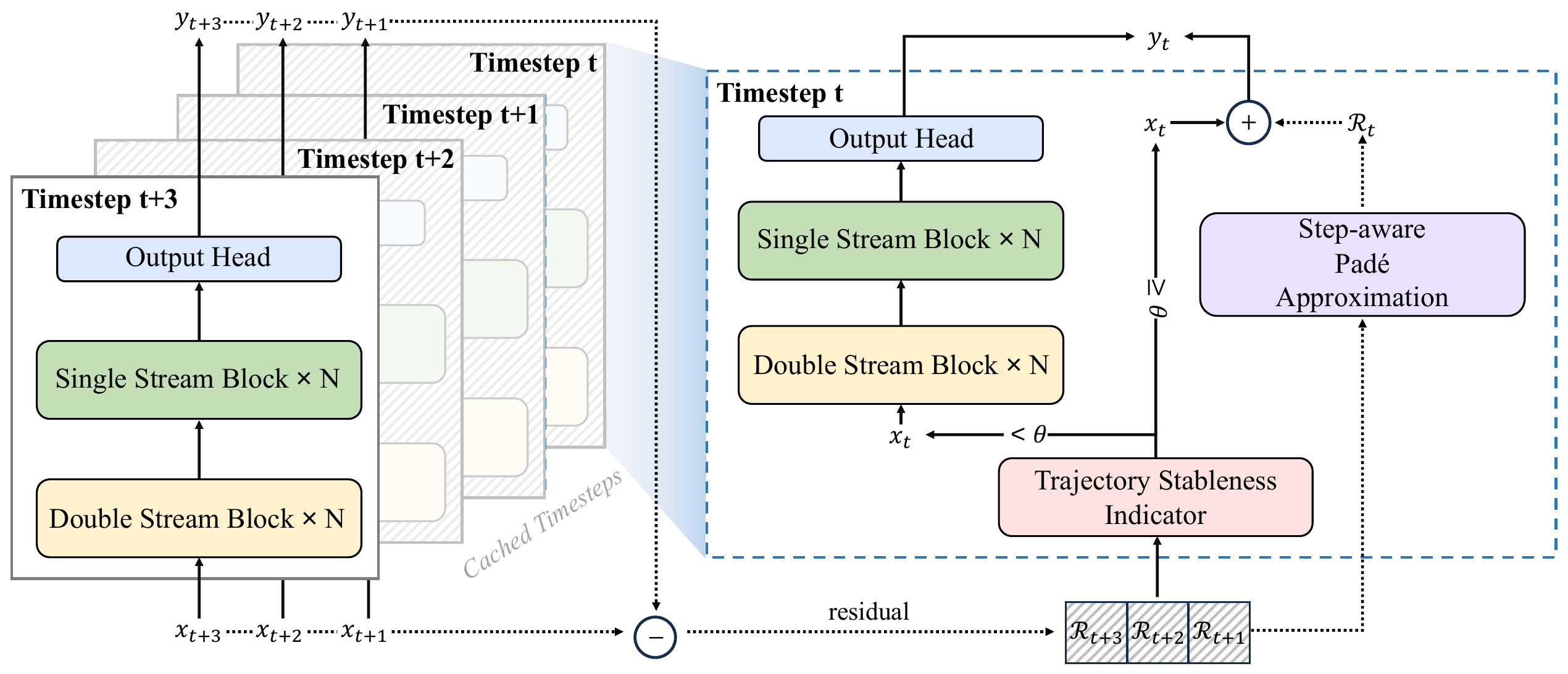}
\caption{
\textbf{Overview of TC-Padé within a cache interval $\mathcal{N}$.} Here, $\mathcal{N}=4$. $\{x_{t+3},x_{t+2},x_{t+1},x_{t}\}$ and $\{y_{t+3},y_{t+2},y_{t+1},y_{t}\}$ denote the input and output of each timestep, respectively. $\{\mathcal{R}_{t+3},\mathcal{R}_{t+2},\mathcal{R}_{t+1},\mathcal{R}_{t}\}$ are the cached residuals. $\theta$ is a predefined threshold. In each cache interval, only the initial timestep performs full computation, while subsequent timesteps adaptively determine their computation mode via the Trajectory Stableness Indicator (TSI).
}
\label{fig:overview}
\end{figure*}

\begin{figure}
\centering
\includegraphics[width=0.5\textwidth]{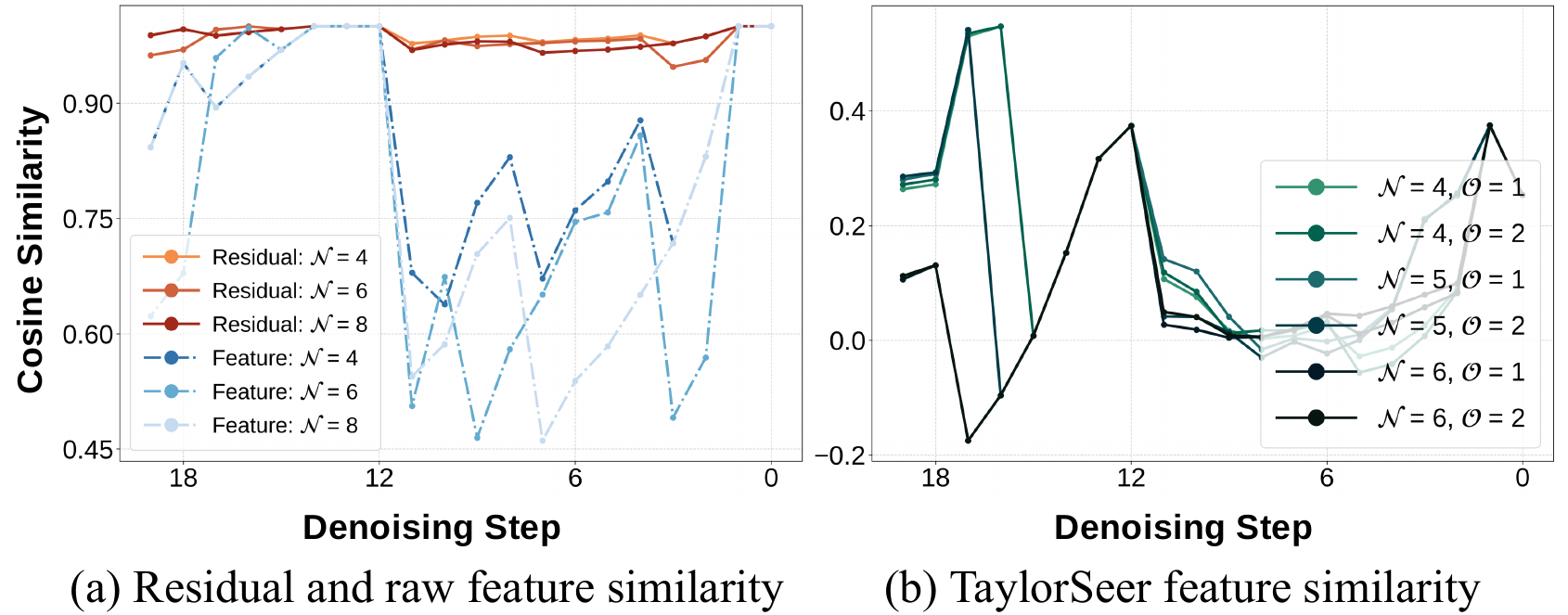}
\caption{
(a) Residual and raw feature similarity between our method and the original sampling schedule, showing that residuals have consistently higher similarity. (b) Raw feature similarity between the original sampling schedule and TaylorSeer under different settings.
}
\label{fig:residual_feature}
\end{figure}

\noindent \textbf{Overview.} 
In this section, we propose TC-Padé, a novel trajectory-consistent residual prediction framework inspired by Padé approximation. As shown in 
Figure~\ref{fig:overview}, our method partitions the sampling trajectory 
into cache intervals of $\mathcal{N}$ timesteps each, employing an adaptive 
computation strategy within each interval.

Specifically, the first timestep performs full computation to establish a reference state. Subsequent timesteps adaptively determine their computation mode via the defined Trajectory Stableness Indicator (TSI) module:
\begin{align}
    \text{TSI}(\mathcal{R}_{t+3}, \mathcal{R}_{t+2}, \mathcal{R}_{t+1}) = \frac{1}{2}\left\|\mathbf{u}_{t+1} - \mathbf{u}_{t+2}\right\|_2,
\end{align}
where $\mathcal{R}_t$ is the residual at timestep t (see Section~\ref{sec:residual}) and $\mathbf{u}_t = (\mathcal{R}_t - \mathcal{R}_{t+1})/\|\mathcal{R}_t - \mathcal{R}_{t+1}\|_2$ denotes the normalized residual difference vector between consecutive timesteps. When $\text{TSI} \geq \theta$, with $\theta$ being a predefined stability threshold, indicating a stable and smooth trajectory, computation is skipped and residuals are predicted via the method introduced in Section~\ref{sec:pade_predict}. Conversely, when $\text{TSI} < \theta$, signaling trajectory instability, full computation is executed to preserve generation fidelity. This adaptive mechanism enables high-quality sampling under large timestep intervals by concentrating computational resources on unstable trajectory regions while accelerating through smooth segments.

\subsection{Padé-Inspired Residual Prediction}
\label{sec:pade_predict}
While Taylor series expansion provides excellent polynomial approximation through derivatives, rational function approximations can better capture certain nonlinear dynamics with fewer historical points~\cite{brezinski1996extrapolation}. In this section, we extend the forecasting paradigm by introducing a Padé-inspired approximation method that operates on residual representations rather than direct features.

\noindent \textbf{Mathematical Foundation: Padé Approximation.}
The Padé approximant is a rational function that approximates a given function using the ratio of two polynomials. Formally, the $\left[m/n\right]$ Padé approximant is defined as:
\begin{align}
    \mathcal{P}_{\left[m/n\right]}=\frac{P_m(x)}{Q_n(x)}=\frac{\sum^m_{i=0}a_ix^i}{1+\sum^n_{j=1}b_jx^j}
\end{align}
where $P_m(x)$ is a polynomial of degree $m$ (numerator) and $Q_n(x)$ is a polynomial of degree $n$ (denominator) with $Q_n(0)=1$ for uniqueness. The key advantage of Padé approximation over pure Taylor expansion lies in its ability to model functions with poles, asymptotic behavior, or rapid nonlinear transitions using rational functions, often achieving faster convergence with fewer terms.

\noindent \textbf{Residual-Based Feature Representation.}
\label{sec:residual}
In modern diffusion transformer, each attention layer $l$ at timestep $t$ processes input features $x^l_t$ through a double-stream or single-stream block. We define the residual between layer $l$ and layer $r$ at timestep $t$ as:
\begin{align}
    \mathcal{R}^{l:r}_t = x^r_t - x^l_t
\end{align}
This residual $\mathcal{R}$ represents the incremental update applied by the layers from $l$ to $r$, capturing the essential transformation independent of the absolute feature values. In practice, we find that caching residuals across the entire DiT block sequence yields optimal performance (see ablation study in Section~\ref{sec:ablation}). For notational simplicity, we omit the layer superscripts in subsequent discussions. Empirically, we observe that \textit{residuals exhibit significantly higher temporal similarity compared to raw features}. In Figure~\ref{fig:residual_feature}(a), we provide robust evidence to support this observation. We compared residual and raw feature similarity between our method and the original sampling schedule. Across the entire sampling steps, residuals have consistently higher cosine similarity than raw features. Previous caching methods like TaylorSeer directly predict raw feature $x_t$. However, as the timestep gap increases, the accumulation of absolute feature change leads to exponential similarity decay. In Figure~\ref{fig:residual_feature}(b), the raw feature similarity between TaylorSeer and the original sampling schedule remains low (less than $0.5$), demonstrating the superiority of residual-based prediction.

\noindent \textbf{Padé-Inspired Feature Forecasting.}
Following the Padé framework, we construct a rational predictor for residuals at timestep $t$ using cached residuals from previous full activation steps $\{t+k, t+k-1,...,t+1\}$. The corresponding resiual history is denoted as $\{\mathcal{R}_{t+k}, \mathcal{R}_{t+k-1},...,\mathcal{R}_{t+1}\}$. The prediction takes the form:
\begin{align}
    \bar{\mathcal{R}}_{t}=\frac{\sum_{i=0}^{m}b_i\mathcal{R}_{t+k-i}}{1+\sum_{j=m+1}^{k-1}a_{j-m}\mathcal{R}_{t+k-j}}
\end{align}
where $m \in [0, k-2]$ is the order of the numerator, and the coefficients $\{b_i\}_{i=0}^m$ and $\{a_j\}_{j=m+1}^{k-1}$ are adaptively determined based on the stability characteristics of the residual trajectory.

The orders $k$ and $m$ control the number of cached historical residuals and the complexity of the rational approximation, respectively. While larger 
$k$ improves prediction accuracy by leveraging more history, it incurs higher memory costs and computational overhead. To strike a balance between expressiveness and computational efficiency, we adopt a low-order Padé-like approximation $\left[2/1\right]$ with $k=3$ and $m=1$:
\begin{align}
    \mathcal{R}_{Pad\acute{e},t}=\frac{b_0\mathcal{R}_{t+3}+b_1\mathcal{R}_{t+2}}{1+a_1\mathcal{R}_{t+1}}
    \label{eq:R_bar}
\end{align}
Once the residual is predicted via Eq.\eqref{eq:R_bar}, the output feature at timestep $t$ is reconstructed as:
\begin{align}
    \bar{x}_{t}=x_{t+1}+\mathcal{R}_{Pad\acute{e},t}
\end{align}
This formulation decouples residual prediction from raw feature prediction, circumventing the need to model the full high-dimensional feature space and allowing the Padé approximation to focus on the more predictable and structured residual dynamics. 

\noindent \textbf{Adaptive Coefficient Design.}
Unlike classical Padé approximation where coefficients are derived analytically from Taylor series matching conditions, the discrete and stochastic nature of diffusion residual trajectories necessitates adaptive, data-driven coefficient design. To prevent the unstable transition from historical caches to current residual, we modulate coefficients by a stability factor that measures the relative magnitude of recent changes:
\begin{align}
    \sigma_{stab}=exp(-\lambda\frac{\lVert \mathcal{R}_{t+1}-\mathcal{R}_{t+2} \rVert}{\lVert \mathcal{R}_{t+1}+\mathcal{R}_{t+2} \rVert})
    \label{eq:stability_factor}
\end{align}
where $\lambda$ is a sensitive parameter. By setting $\lambda$ to a relatively large value, this exponential decay ensures that $\sigma_{stab} \to 0$ when residuals change rapidly, and $\sigma_{stab} \to 1$ when residuals remain stable. The coefficients are then defined as:
\begin{align}
    b_0=2\sigma_{stab}, \quad b_1=-\sigma_{stab}, \quad a_1=\frac{1}{\lambda}\sigma_{stab}
\end{align}
The coefficient formulations are designed to ensure numerical stability across diverse diffusion schedules. Detailed discussions on parameter sensitivity and design rationale are provided in the Appendix.

\subsection{Denoising Step-aware Strategy}
The effectiveness of feature caching varies substantially across the timesteps of diffusion denoising. Early stages (high noise) involve rapid, large-scale structural formation, whereas late stages (low noise) focus on fine-grained detail refinement~\cite{ma2024deepcache}. We partition the denoising process into three phases, each associated with different residual update strategies, which leads to the final residual prediction target as:
\begin{align}
    \bar{\mathcal{R}}_t = 
    \begin{cases}
        \alpha_1\mathcal{R}_{t+1}+\alpha_2\mathcal{R}_{t+2}, & t > 0.7T \\
        \mathcal{R}_{Pad\acute{e},t}, & 0.2T \le t \le 0.7T \\
        \mathcal{R}_{Pad\acute{e},t}+\beta(\mathcal{R}_{t+1}-\mathcal{R}_{t+2}), & t < 0.2T
    \end{cases}
\end{align}
where $T$ denotes the total number of denoising steps. In the early phase where structural information rapidly evolves, we directly use a weighted combination of the two most recent residuals with $\alpha_1+\alpha_2=1$. For the middle phase, we utilize the full Padé approximant to exploit long-range dependencies in the residual trajectory. Finally, in the refinement phase, we augment the Padé prediction with a first-order difference term $\beta(\mathcal{R}_{t+1}-\mathcal{R}_{t+2})$, where $\beta$ is a small coefficient that captures subtle velocity changes in residual evolution. This step-aware strategy ensures that our caching mechanism maintains both computational efficiency and generation quality across the entire denoising trajectory.

\section{Experiments}

\begin{table*}[ht]
\centering
\caption{
Quantitative results for \textbf{text-to-image} generation on COCO 2017 dataset. The best results are in \textbf{bold}, and the second best are \underline{underlined}. Values marked with $\dagger$ indicate severe degradation in output image quality, with results falling outside the acceptable range for meaningful comparison. Additional metrics are provided in the Appendix.
}
\label{tab:text-to-image}
\begin{adjustbox}{max width=\textwidth}
\begin{tabular}{c c c c c c c c c}
\toprule
\multirow{2}{*}{\begin{tabular}[c]{@{}c@{}}Cache Type\end{tabular}} & \multirow{2}{*}{Method} & \multicolumn{2}{c}{Efficiency} & \multicolumn{4}{c}{Visual Quality} \\
\cmidrule(lr){3-4} \cmidrule(lr){5-9}
& & Latency (s)$\downarrow$ & FLOPs (T)$\downarrow$ & FID$\downarrow$ & CLIP Score$\uparrow$ & PSNR$\uparrow$ & SSIM$\uparrow$ & LPIPS$\downarrow$\\
\midrule
- & Flux.1-dev, 20 steps & 9.22(1.00$\times$)  &1487.80(1.00$\times$) &  23.38 & 32.10 & - & - & - \\
\midrule

\multirow{4}{*}{\begin{tabular}[c]{@{}c@{}} Reuse-based \end{tabular}} & ToCa ($\mathcal{N}=5$) &5.09(1.81$\times$) &509.48(2.92$\times$) &24.18 & 31.48 & 17.291 & 0.6132 & 0.4805 \\
& $\Delta$-DiT ($\mathcal{N}=3$) &5.13(1.80$\times$) &694.54(2.14$\times$) &24.03 &31.00 & 16.234 & 0.5773 & 0.5333 \\
& TeaCache (slow) & 6.49(1.42$\times$) & 982.45(1.51$\times$) & \underline{23.90} & 31.38 & 18.890 & 0.7241 & 0.3500 \\
& TeaCache (fast) & 4.29(2.15$\times$) & 610.60(2.44$\times$) & 24.11 &31.50 & 18.016 & 0.6898 & 0.4192 \\
\cmidrule(lr){2-9}
\multirow{4}{*}{\begin{tabular}[c]{@{}c@{}} Prediction-based \end{tabular}} & TaylorSeer ($\mathcal{N}=5$, $\mathcal{O}=2$) & 3.97(2.31$\times$) &\underline{461.96(3.22$\times$)} & $\dagger$ & 31.52 & 17.464 & 0.5248 & 0.6163  \\
& TaylorSeer ($\mathcal{N}=6$, $\mathcal{O}=2$) &\underline{3.56(2.59$\times$)} & \textbf{387.59(3.84$\times$)} & $\dagger$&30.95 & 16.567 & 0.5625 & 0.5449  \\
& \cellcolor{gray!15} TC-Padé (slow) & \cellcolor{gray!15} 4.20(2.20$\times$) & \cellcolor{gray!15} 582.60(2.55$\times$) & \cellcolor{gray!15} \textbf{23.85} & \cellcolor{gray!15} \textbf{31.90} & \cellcolor{gray!15} \textbf{24.673} & \cellcolor{gray!15} \textbf{0.8607} & \cellcolor{gray!15} \textbf{0.1435} \\
& \cellcolor{gray!15} TC-Padé (fast) & \cellcolor{gray!15} \textbf{3.20(2.88$\times$)} & \cellcolor{gray!15} 506.23(2.94$\times$) & \cellcolor{gray!15} 24.14 & \cellcolor{gray!15} \underline{31.82} & \cellcolor{gray!15} \underline{21.962} & \cellcolor{gray!15} \underline{0.7823} & \cellcolor{gray!15} \underline{0.2896} \\
\bottomrule
\end{tabular}
\end{adjustbox}
\end{table*}

\begin{table*}[ht]
\centering
\caption{Quantitative comparison for \textbf{text-to-video} generation for Wan2.1-1.3B on VBench-2.0.}
\label{tab:text-to-video}
\begin{adjustbox}{max width=\textwidth}
\begin{tabular}{@{}cccccccc@{}}
\toprule
\multirow{2}{*}{Cache Type} &
\multirow{2}{*}{Method} 
  & \multicolumn{2}{c}{Efficiency} 
  & \multicolumn{4}{c}{Visual Quality} \\
\cmidrule(lr){3-4} \cmidrule(lr){5-8}
& & Latency (s)$\downarrow$ & FLOPs (T)$\downarrow$ & VBench-2.0$\uparrow$ & PSNR$\uparrow$ & SSIM$\uparrow$ & LPIPS$\downarrow$ \\
\midrule
- & Wan2.1, 20 steps & 88(1.00$\times$) & 3568.83(1.00$\times$) & 64.16\% & - & -& - \\ \midrule
\multirow{2}{*}{Reuse-based} & Teacache (slow) & 75(1.17$\times$) & 3076.39(1.16$\times$) & \textbf{60.73\%} & \textbf{27.19} & \textbf{0.8674} & \textbf{0.1073} \\
& Teacache (fast) & 61(1.44$\times$) & 2583.95(1.28$\times$) &58.40\% & 21.35 & 0.6142 & 0.3265 \\ \cmidrule(lr){2-8}
\multirow{3}{*}{Prediction-based} & TaylorSeer ($\mathcal{N}=3$, $\mathcal{O}=1$) & 67(1.31$\times$) & \underline{1954.23(1.83$\times$)} & 54.74\% & 15.02 & 0.3596 & 0.5657 \\
& TaylorSeer ($\mathcal{N}=4$, $\mathcal{O}=1$) & \underline{53(1.66$\times$)} & \textbf{1876.24(1.90$\times$)} & 54.50\% & 14.93 & 0.3532 & 0.5856 \\
& \cellcolor{gray!15} TC-Padé (fast) & \cellcolor{gray!15} \textbf{51(1.72$\times$)} & \cellcolor{gray!15} 2055.24(1.74$\times$) & \cellcolor{gray!15} \underline{60.38\%} & \cellcolor{gray!15} \underline{21.70} & \cellcolor{gray!15} \underline{0.6387} & \cellcolor{gray!15} \underline{0.3001} \\
\bottomrule
\end{tabular}
\end{adjustbox}
\end{table*}

\begin{table*}[ht]
\centering
\caption{Quantitative comparison for class-conditional image generation on ImageNet with DiT-XL/2.}
\label{tab:class-to-image}
\begin{adjustbox}{max width=\textwidth}
\begin{tabular}{@{}cccccccc@{}}
\toprule
\multirow{2}{*}{Cache Type} & \multicolumn{1}{c}{\multirow{2}{*}{Method}} & \multicolumn{2}{c}{Efficiency} & \multicolumn{4}{c}{Visual Quality} \\
\cmidrule(lr){3-4} \cmidrule(lr){5-8}
& & Latency (s)$\downarrow$ & FLOPs (T)$\downarrow$ & FID$\downarrow$ & IS$\uparrow$ & Precision$\uparrow$ & Recall$\uparrow$ \\
\midrule
- & DiT-XL/2, 20steps & 1.71(1.00$\times$) & 9.49(1.00$\times$) & 3.56 & 221.27 & 0.78 & 0.58 \\
\midrule
\multirow{2}{*}{Reuse-based} & ToCa ($\mathcal{N}=3$) & 1.26(1.35$\times$) & 4.01(2.37$\times$) & 10.72 & 164.40 & 0.69 & 0.49 \\
& $\Delta$-DiT ($\mathcal{N}=3$) & 1.31(1.31$\times$) & 6.43(1.48$\times$) & 8.86 & 170.96 & 0.70 & 0.55 \\
\cmidrule(lr){2-8}
\multirow{3}{*}{Prediction-based} & TaylorSeer ($\mathcal{N}=4$, $\mathcal{O}=3$)  & \textbf{1.13(1.51$\times$}) &\textbf{2.85(3.32$\times$)} & 7.86 & 175.11 & \underline{0.71} & \underline{0.53}  \\
& TaylorSeer ($\mathcal{N}=3$, $\mathcal{O}=2$) & 1.19(1.44$\times$) & \underline{3.80(2.49$\times$)} & \underline{7.84} & \underline{175.99} & \underline{0.71} & \underline{0.53} \\
& \cellcolor{gray!15} TC-Padé (fast) & \cellcolor{gray!15} \underline{1.17(1.46$\times$)} & \cellcolor{gray!15} 5.79(1.64$\times$) & \cellcolor{gray!15} \textbf{6.93} & \cellcolor{gray!15} \textbf{185.12} & \cellcolor{gray!15} \textbf{0.72} & \cellcolor{gray!15} \textbf{0.54} \\
\bottomrule
\end{tabular}
\end{adjustbox}
\end{table*}
\subsection{Experimental Setup}
\noindent \textbf{Base Models.}
To demonstrate the effectiveness of our method, we apply our acceleration
technique to three state-of-the-art visual generative models: the text-to-image generation model FLUX.1-dev~\cite{flux2024}, text-to-video generation model Wan2.1-1.3B~\cite{wan2025wan}, and the class-conditional image generation model DiT-XL/2~\cite{peebles2022scalable}.

\noindent \textbf{Evaluation Metrics.}
We evaluate our proposed method from along two dimensions: inference efficiency and visual quality. For evaluating inference efficiency, we use Floating Point Operations (FLOPs) and inference latency as metrics. Visual quality is measured across three tasks. For text-to-image generation, we randomly sample 50,000 prompts from the COCO 2017~\cite{lin2014microsoft} training set and report Fréchet Inception Distance (FID)~\cite{heusel2017gans} for overall distributional fidelity, together with four text–image alignment and perceptual metrics: CLIP Score~\cite{radford2021learning}, PickScore~\cite{kirstain2023pick}, Aesthetic Score~\cite{schuhmann2022laion}, and ImageReward~\cite{xu2023imagereward}. For text-to-video generation, we adopt VBench-2.0~\cite{zheng2025vbench} as the evaluation benchmark, which provides a comprehensive suite of video quality assessments. Additionally, we report PSNR, SSIM, and LPIPS~\cite{zhang2018unreasonable} to quantitatively measure the pixel-level and perceptual similarity of the generated videos. For class-conditional image generation, we uniformly sample all 1,000 ImageNet~\cite{russakovsky2014imagenet} classes and generate 50 images per class at 256×256 resolution, using FID-50k as the primary metric and additionally reporting Inception Score (IS), Precision, and Recall.

\noindent \textbf{Implementation Detail.}
All experiments are carried out on the NVIDIA L40 GPUs with Pytorch. We fix the number of denoising steps to 20 for all models. The sensitive parameter $\lambda$ in Eq.\eqref{eq:stability_factor} is set to 10. We provide two configurations of our method: TC-Padé (slow), which uses a TSI threshold $\theta=1.0$, and TC-Padé (fast), which uses $\theta=0.7$. All reported results are obtained under these settings unless otherwise noted.

\subsection{Main Results}
\subsubsection{Text-to-Image Generation}
\begin{figure}
\centering
\includegraphics[width=0.47\textwidth]{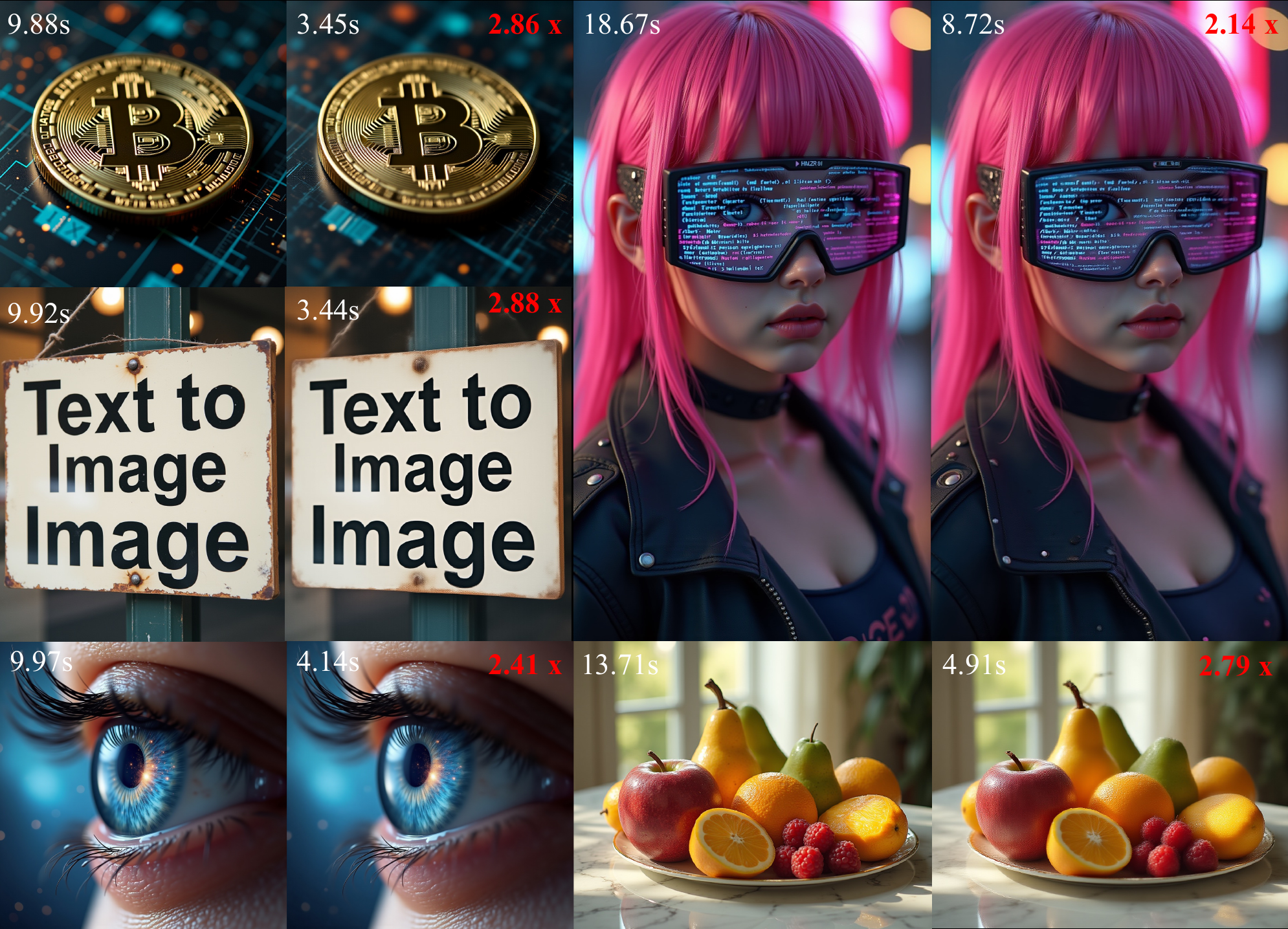}
\caption{
\textbf{Qualitative results of FLUX.1-dev across multiple resolutions.} Each image pair shows the original 20-step sampling result and TC-Padé accelerated sampling result. Numbers indicate latency (top-left) and speedup ratio (top-right).
}
\label{fig:qualitative_example}
\end{figure}

In this subsection, we evaluate our method on the COCO 2017 training set, with results summarized in Table~\ref{tab:text-to-image}. We compare against ToCa~\citep{zou2024accelerating}, $\Delta$-DiT~\cite{chen2024delta}, TeaCache~\cite{liu2025timestep}, and TaylorSeers~\cite{liu2025reusing}. To ensure fairness, we adopt their publicly available implementations and select the optimal hyperparameters as reported in their respective works. While existing caching-based approaches achieve acceleration ratios ranging from 1.81$\times$ to 2.59$\times$, they exhibit significant visual quality degradation under limited sampling steps. In contrast, TC-Padé demonstrates consistently superior performance across both sampling efficiency and visual quality. Notably, TC-Padé (fast) achieves a $2.88\times$ speedup with only 20 sampling steps, while maintaining high fidelity with PSNR of 21.962, SSIM of 0.7823, and LPIPS of 0.2896—substantially outperforming existing methods in both pixel-level accuracy and perceptual quality. Qualitative examples in Figure~\ref{fig:qualitative_example} highlights TC-Padé’s superior ability to generate highly consistent images at high speedup ratio.

\subsubsection{Text-to-Video Generation}
For text-to-video generation on Wan2.1-1.3B, our method achieves the best efficiency-quality trade-off. As shown in Table~\ref{tab:text-to-video}, TC-Padé (fast) delivers a 1.72$\times$ latency speedup and 1.74$\times$ FLOPs reduction while maintaining a VBench-2.0 score of 60.38\%—only 3.78 points below the 20-step baseline (64.16\%). Moreover, it preserves strong reconstruction fidelity with PSNR of 21.70, SSIM of 0.6387, and LPIPS of 0.3001, demonstrating well-balanced performance across both perceptual and pixel-level quality metrics.

In contrast, existing methods exhibit suboptimal trade-offs between efficiency and quality. Reuse-based approaches like Teacache either achieve minimal acceleration with preserved quality or provide moderate speedup at the cost of noticeable degradation. Prediction-based TaylorSeer methods, while offering higher acceleration ratios (up to 1.90$\times$), deviate drastically from the original sampling results with substantial degradation in PSNR, SSIM and LPIPS metrics. Overall, TC-Padé attains the largest practical speedup ($>$1.7$\times$) with minimal quality loss.

\subsubsection{Class-conditional Image Generation}
To evaluate the effectiveness of our method on class-conditional image generation tasks, we conduct experiments on ImageNet 256×256 using DiT-XL/2 with 20 sampling steps. We compare our approach against representative reuse-based methods (ToCa and $\Delta$-DiT) and prediction-based methods (TaylorSeer with different configurations), as well as the baseline DiT-XL/2 without acceleration.
Table~\ref{tab:class-to-image} presents the quantitative results. Our TC-Padé method achieves the best generation quality with FID-50k of 6.93 and IS of 185.12 while delivering 1.46$\times$ latency speedup and 1.64$\times$ FLOPs reduction. In comparison, reuse-based methods suffer from significant quality degradation: ToCa exhibits severe performance drop with FID of 10.72, while $\Delta$-DiT shows FID of 8.86 with only 1.31$\times$ acceleration. Among prediction-based methods, TaylorSeer configurations achieve higher computational savings (up to 3.32$\times$ FLOPs reduction) but at the cost of quality degradation. Notably, our method maintains superior Precision (0.72) and Recall (0.54) scores, indicating better balance between sample fidelity and diversity.

\subsection{Ablation Study}
\label{sec:ablation}
Block-level residual caching provides the best quality--efficiency trade-off, while the TSI threshold smoothly controls the balance between conservative prediction and speed. Full residual and threshold ablations, qualitative comparisons, and quantized deployment results are provided in Appendix~\ref{sec:supp_extended}.

\section{Conclusion}
In this work, we introduce TC-Padé, a trajectory-consistent feature prediction framework that addresses the limitations of existing caching methods in practical low-step diffusion sampling. By leveraging Padé approximation on residual representations, TC-Padé effectively captures nonlinear, phase-dependent feature dynamics that polynomial methods cannot represent. Through adaptive coefficient modulation and step-aware prediction strategies, our method maintains trajectory consistency under large timestep intervals. Experiments across image and video generation demonstrate strong quality--efficiency trade-offs, supporting latency-sensitive deployment.


{
    \small
    \bibliographystyle{ieeenat_fullname}
    \bibliography{main}
}
\clearpage
\setcounter{page}{1}
\maketitlesupplementary


\section{Extended Ablations, Qualitative Results, and Deployment}
\label{sec:supp_extended}
This section collects secondary analyses moved from the main paper to satisfy the eight-page main-text limit.

\begin{figure*}[t]
    \centering
    \parbox{0.48\textwidth}{
        \centering
        \small
        \captionof{table}{Ablation study on the impact of cached residual granularity. Entire block level caching achieves optimal performance-efficiency trade-off.}
        \begin{tabular}{@{}c|c|cccc@{}}
        \toprule
        Granularity & Latency (s) & Aes$\uparrow$ & CLIP$\uparrow$ & ImgRwd$\uparrow$ \\
        \midrule
        Double-stream & 6.78(1.36$\times$) & 5.10 & 31.31 & 0.7921 \\
        Single-stream & 4.75(1.94$\times$) & 5.69 & 31.66 & 0.8717 \\
        Entire Block & \textbf{3.20(2.88$\times$)} & \textbf{5.76} & \textbf{31.83} & \textbf{0.9184} \\
        \bottomrule
        \end{tabular}%
        \label{tab:residual_granularity}

        \vspace{1em} 

        \captionof{table}{Ablation study on the impact of TSI threshold $\theta$.}
        \begin{tabular}{@{}c|c|cccc@{}}
        \toprule
        $\theta$ & Latency (s) & Aes$\uparrow$ & CLIP$\uparrow$ & ImgRwd$\uparrow$ \\
        \midrule
        1.3 & 5.66(1.63$\times$) & \textbf{5.80} & \textbf{32.02} & \textbf{0.9562} \\
        1.0 & 4.20(2.20$\times$) & 5.77 & 31.97 & 0.9236 \\
        0.7 & \textbf{3.20(2.88$\times$)} & 5.76 & 31.83 & 0.9184 \\
        \bottomrule
        \end{tabular}
        \label{tab:tsi_threshold}

        \vspace{1em}

        \captionof{table}{Performance comparison between FLUX.1-dev, TC-Padé and TC-Padé with quantization.}
        \begin{tabular}{@{}c|cccc@{}}
        \toprule
         & FID$\downarrow$ & CLIP$\uparrow$ & Aes$\uparrow$ \\
        \midrule
        FLUX.1-dev & 23.38 & 32.10 & 6.25 \\
        TC-Padé & 24.14 & 31.82 & 6.11 \\
        TC-Padé+Quant & 24.31 & 31.08 & 6.01 \\
        \bottomrule
        \end{tabular}
        \label{tab:quantization}
    }
    \hfill 
    \begin{minipage}{0.48\textwidth}
        \centering
        \includegraphics[width=0.9\linewidth]{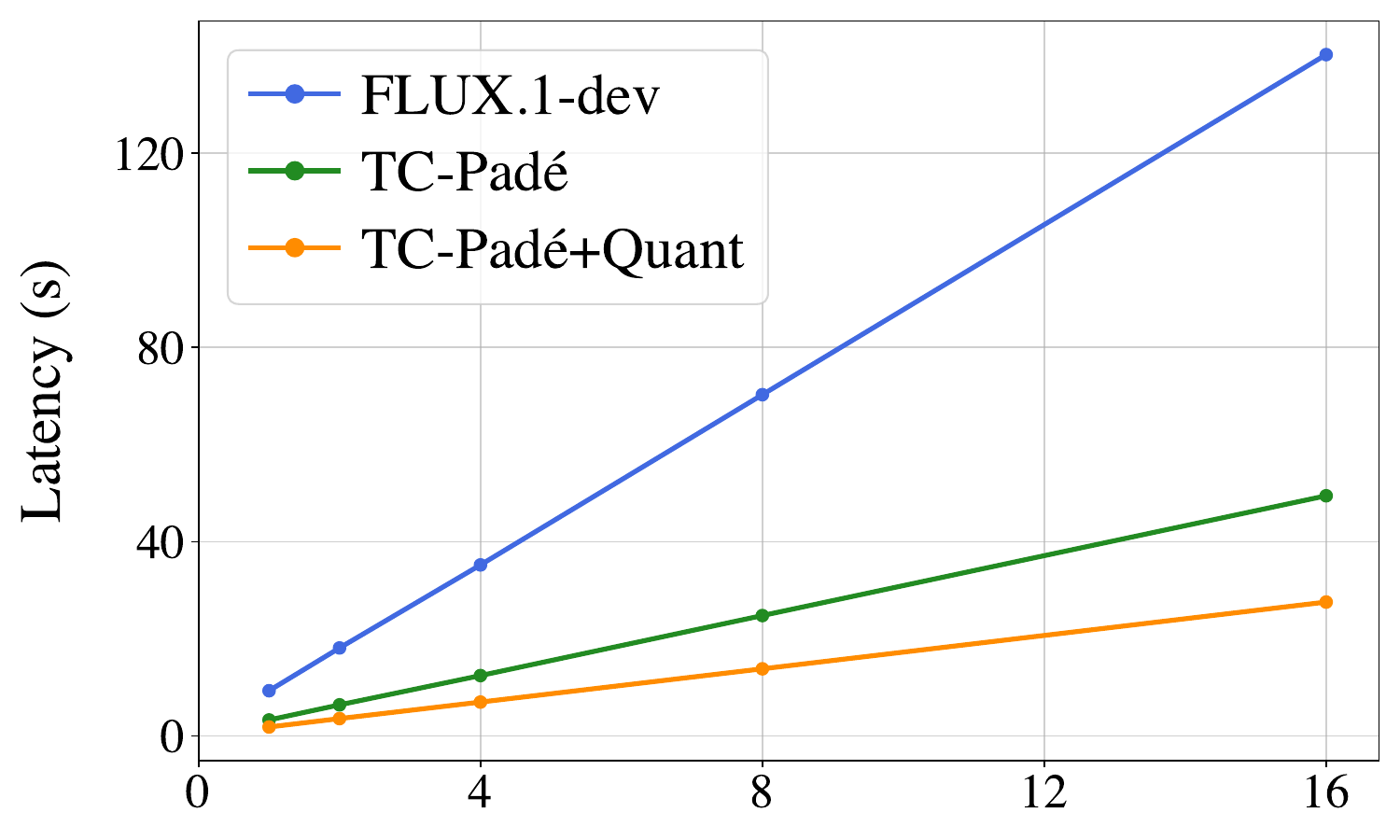}
        
        
        \includegraphics[width=0.9\linewidth]{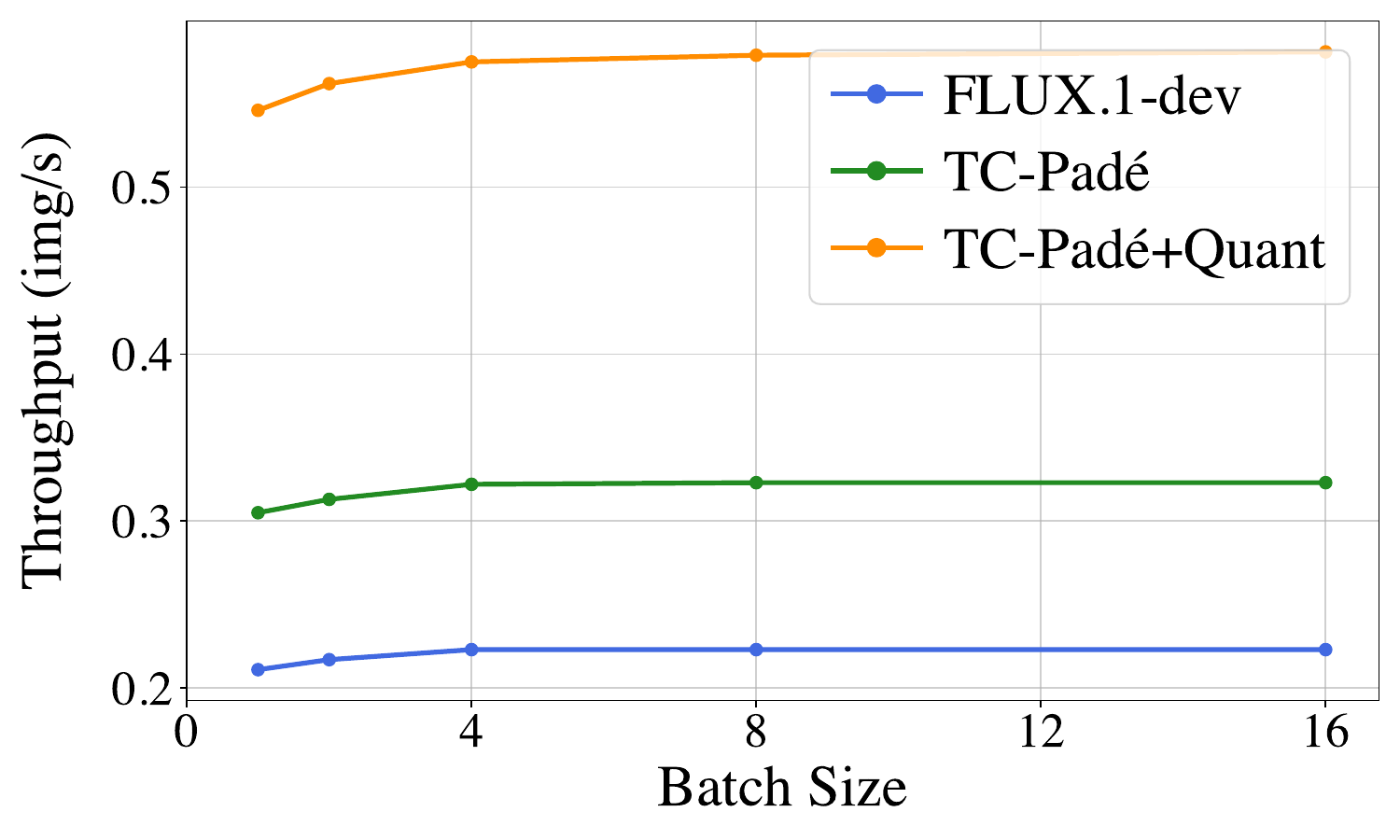}
        
        \caption{TC-Padé with quantization achieves the lowest latency and highest throughput compared to FLUX.1-dev and TC-Padé across different batch sizes.}
        \label{fig:efficiency_compare}
    \end{minipage}
\end{figure*}

\subsection{Multi-resolution Qualitative Results}

\subsection{Ablation Study}
\label{sec:supp_ablation}
\noindent \textbf{Effect of cached residual granularity.}
To determine the optimal granularity for residual caching, we conduct ablation studies on FLUX.1-dev with three different caching strategies while keeping all other hyperparameters constant. Specifically, we compare caching residuals at the granularity of: (1) Double-stream block only, (2) Single-stream block only, and (3) Entire block, which encompasses the complete block including both double-stream and single-stream components. We evaluate all configurations on the DrawBench~\cite{saharia2022photorealistic} benchmark to ensure comprehensive assessment of generation quality. Table~\ref{tab:residual_granularity} presents the quantitative results. Block-level residual caching achieves optimal performance across all metrics, demonstrating a 12.9\% improvement in Aesthetic Score and a 15.9\% improvement in Image Reward compared to double-stream block caching, while providing a 2.88$\times$ speedup.

\noindent \textbf{Effect of TSI threshold.}
To investigate the impact of TSI threshold, we conduct ablation experiments on FLUX.1-dev with three different threshold values: 1.3, 1.0, and 0.7. The TSI threshold $\theta$ is a critical hyperparameter that controls the trade-off between sampling efficiency and generation quality. A higher threshold indicates stricter stability requirements, leading to more conservative skipping behavior, while a lower threshold enables more aggressive acceleration by allowing residual prediction on less stable trajectory segments. Table~\ref{tab:tsi_threshold} presents the results, as $\theta$ decreases from 1.3 to 0.7, the speedup increases from 1.63$\times$ to 2.88$\times$, enabling more aggressive timestep skipping. However, this comes at the cost of minor quality degradation: Image Reward drops from 0.9562 to 0.9184, while Aesthetic and CLIP scores remain relatively stable with marginal decreases. The intermediate setting $\theta=1$ provides a balanced configuration with 2.20$\times$ speedup and competitive quality metrics.

\subsection{Deployment Efficiency}
This subsection evaluates TC-Padé's efficiency by measuring latency and throughput across varying batch sizes. We further demonstrate that our method is orthogonal to other acceleration techniques such as quantization~\cite{xiao2023smoothquant}. As shown in Figure~\ref{fig:efficiency_compare}, TC-Padé with quantization achieves approximately 6$\times$ latency reduction over the original FLUX.1-dev. At batch size 1, TC-Padé with quantization reduces generation time from 9 seconds to 1.83 seconds. This speedup scales consistently: at batch size 16, latency decreases from 140.31 seconds to 27.56 seconds, demonstrating robust scalability.

In terms of generation throughput, FLUX.1-dev maintains a relatively constant throughput of 0.22 img/s regardless of batch size. TC-Padé alone achieves approximately 0.32 img/s, representing a consistent 1.4-1.5$\times$ improvement. The combination with quantization further boosts throughput to 0.54-0.57 img/s, achieving an overall 2.5$\times$ speedup over the baseline. Importantly, as shown in Table~\ref{tab:quantization}, this substantial efficiency gain comes with minimal quality degradation. This demonstrates that our temporal consistency optimization synergizes effectively with orthogonal acceleration techniques.

\section{More Quantitative Results}
\noindent \textbf{Text-to-Image Generation.}
To provide a more comprehensive assessment of our method's performance, we present additional evaluation results on the DrawBench benchmark using Aesthetic Score and ImageReward metrics, as referenced in Tabel~\ref{tab:drawbench_results}.

\begin{table}[ht]
\centering
\caption{
Quantitative results for \textbf{text-to-image} generation on \textbf{DrawBench}. 
Higher is better for quality metrics, and lower is better for efficiency metrics. 
The best results are in \textbf{bold}, and the second best are \underline{underlined}.
}
\begin{tabular}{c c c}
\toprule
\multirow{2}{*}{\begin{tabular}[c]{@{}c@{}}DrawBench\end{tabular}} & \multicolumn{2}{c}{Visual Quality} \\
\cmidrule(lr){2-3}
 & ImgRwd$\uparrow$ & Aesthetic$\uparrow$ \\
\midrule
Flux.1[dev], 20steps & 1.01 & 5.83 \\
\midrule
$\Delta$-DiT (N = 3) & 0.52 & 5.42  \\
ToCa(N=5) & 0.82 & 5.24  \\
TeaCache($\delta=0.25$) & 0.85 & \underline{5.73}  \\
TeaCache($\delta=0.4$)  & \underline{0.86} & 5.71  \\
TeaCache($\delta=0.6$)  & 0.74 & 5.62  \\
TaylorSeer ($\mathcal{N}=5$, $\mathcal{O}=2$) & 0.69 & 4.75 \\
TaylorSeer ($\mathcal{N}=6$, $\mathcal{O}=2$) & 0.68 & 4.40 \\
\cellcolor{gray!15} TC-Padé (fast) & \cellcolor{gray!15} \textbf{0.93} & \cellcolor{gray!15} \textbf{5.83} \\
\bottomrule
\end{tabular}
\label{tab:drawbench_results}
\end{table}

\begin{figure}
\centering
\includegraphics[width=0.47\textwidth]{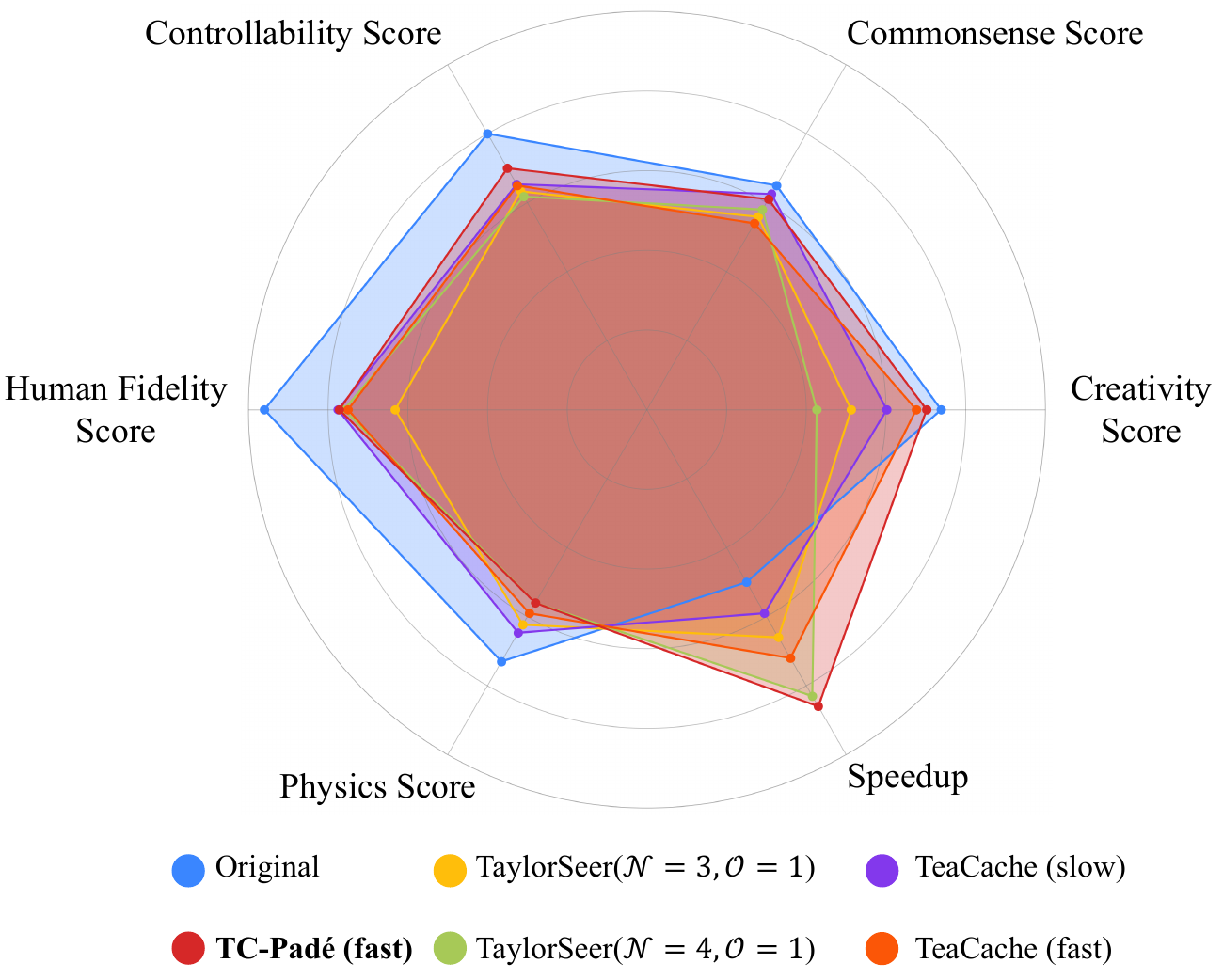}
\caption{
VBench-2.0 metrics and acceleration ratio of TC-Padé and other methods.
}
\label{fig:vbench_score}
\end{figure}

\noindent \textbf{Text-to-Video Generation.}
For text-to-video generation, we provide a more detailed analysis of our method's performance on VBench-2.0 beyond the Total Score reported in the main text. Figure~\ref{fig:vbench_score} presents a radar chart showing the performance across five key dimensions: Creativity Score, Commonsense Score, Controllability Score, Human Fidelity Score, and Physics Score, comparing our method against baseline approaches. Our method demonstrates balanced performance between generation quality and efficiency.

\begin{figure}
\centering
\includegraphics[width=0.47\textwidth]{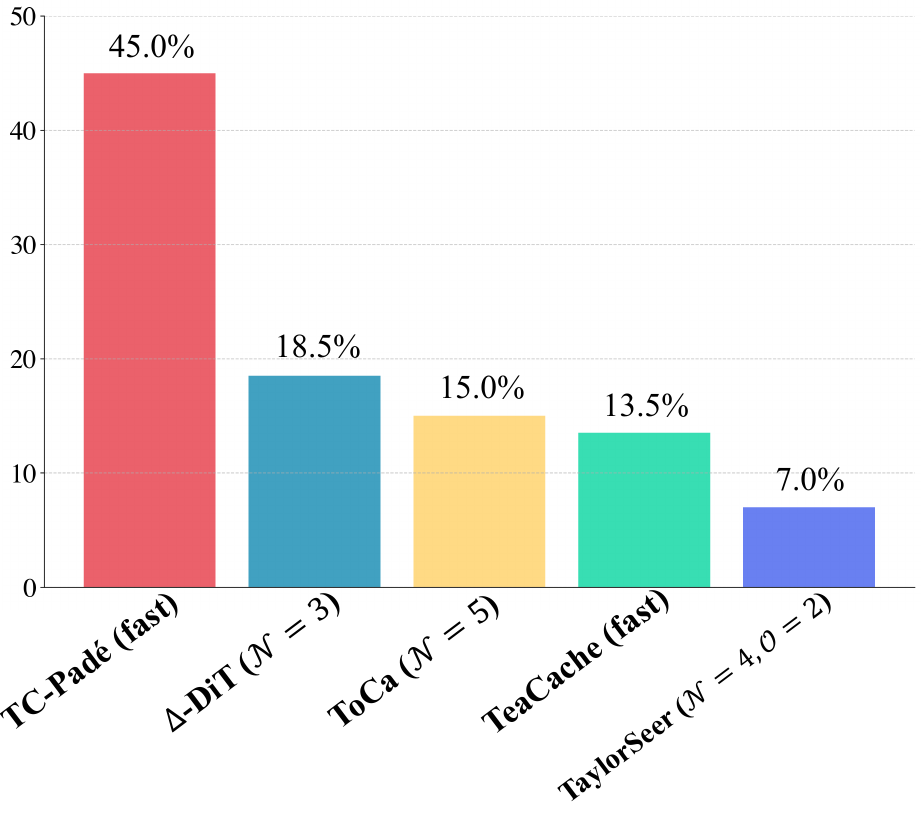}
\caption{
User preference comparison for text-to-image generation on FLUX.1-dev across different caching methods.
}
\label{fig:user_preference}
\end{figure}

\noindent \textbf{User Study.}
We conducted a user study to evaluate text-to-image generation quality on FLUX.1-dev across different caching methods. Fifty volunteers participated in a blind comparison study, where they were presented with generated images alongside their corresponding text prompts and asked to select the method that best preserved image quality. To ensure unbiased evaluation, participants were not informed which method produced each image. The results are presented in Figure~\ref{fig:user_preference}. our method achieves a preference rate of 45\%, indicating superior perceptual quality compared to other caching approaches.

\section{Rationale for Coefficient Design}
\subsection{Stability-Aware Coefficient Analysis}
The application of Padé approximation to diffusion model cache prediction requires careful adaptation from its classical formulation. In traditional Padé approximation, coefficients are analytically derived by matching Taylor series expansions to achieve optimal rational function approximation of smooth, deterministic functions. However, direct application of fixed coefficients leads to two critical failure modes:
\begin{itemize}
\item \textbf{Over-extrapolation}: In regions where residuals change rapidly (early denoising steps), aggressive prediction amplifies errors.
\item \textbf{Under-utilization}: In stable regions (late denoising steps), conservative prediction fails to leverage temporal coherence.
\end{itemize}
To address this, we introduce a stability factor $\sigma_{stab}$ that quantifies local trajectory smoothness and modulates coefficient aggressiveness accordingly. In Eq.~\ref{eq:stability_factor}, the numerator $\lVert \mathcal{R}_{t+1}-\mathcal{R}_{t+2} \rVert$ measures the magnitude of recent residual change, while the denominator $\lVert \mathcal{R}_{t+1}+\mathcal{R}_{t+2} \rVert$ provides scale normalization to ensure consistency across different noise levels. When the trajectory exhibits rapid variation (large numerator), the exponential term $exp(-\lambda\cdot)$ drives $\sigma_{stab} \to 0$, effectively reducing reliance on potentially misleading historical patterns. Conversely, when residuals evolve smoothly, $\sigma_{stab} \to 1$ enables full utilization of temporal correlations for accurate prediction. $\sigma_{stab} \in (0,1)$ ensures numerical stability and prevents coefficient explosion.

The coefficients $b_0$ and $b_1$ maintain a fixed ratio of $2:-1$, which implements a second-order finite difference approximation that is optimal for smooth trajectory segments when $\sigma_{stab} \approx 1$. The denominator coefficient $a_1=\frac{1}{\lambda}\sigma_{stab}$ scales inversely with $\lambda$. Since $\lambda$ is set to a large value (e.g., $\lambda \geq 10$, this ensures $a_1$ remains small, preventing division instabilities while still allowing adaptive normalization. Since all coefficients scale linearly with $\sigma_{stab}$, the formulation guarantees smooth transitions across different stability regimes and avoids discontinuous prediction behavior at regime boundaries.
\subsection{Parameter Sensitivity Analysis}
\begin{table}[ht]
\centering
\caption{Impact of sensitivity parameter $\lambda$ on generation quality.}
\begin{adjustbox}{max width=0.48\textwidth}
\begin{tabular}{@{}c|cccccc@{}}
\toprule
\multirow{2}{*}{\begin{tabular}[c]{@{}c@{}}$\lambda$\end{tabular}} & \multicolumn{6}{c}{Visual Quality} \\
\cmidrule(lr){2-7}
 & Aesthetic$\uparrow$ & PickScore$\uparrow$ & ImgRwd$\uparrow$ & PSNR$\uparrow$ & SSIM$\uparrow$ & LPIPS$\downarrow$ \\
\midrule
5 & 5.79 & 22.48 & 0.866 & 23.850 & 0.8080 & 0.2929 \\
10 & \textbf{5.83} & \textbf{22.64} & \textbf{0.9378} & \textbf{24.436} & \textbf{0.8376} & \textbf{0.2463} \\
15 & 5.79 & 22.43 & 0.8408 & 23.752 & 0.7981 & 0.3087 \\
\bottomrule
\end{tabular}
\label{tab:lambda}
\end{adjustbox}
\end{table}
The sensitivity parameter $\lambda$ controls the sharpness of the stable-to-unstable transition.
\begin{itemize}
\item \textbf{When $\lambda$ is small (1-5)}, the stability factor transitions gradually between regimes, but this fails to sufficiently suppress predictions during unstable phases, leading to potential extrapolation errors.
\item \textbf{Medium values $\lambda$ (5-15)} provide balanced sensitivity that sharply distinguishes stable from unstable trajectories while maintaining smooth coefficient adaptation.
\item \textbf{When $\lambda$ exceeds 15}, the exponential decay becomes extremely sharp, which may cause abrupt switching between prediction modes and introduce discontinuity artifacts in the generated samples.
\end{itemize}
We evaluate different $\lambda$ values on DrawBench prompts. Results in Tabel~\ref{tab:lambda} show that $\lambda=10$ achieves the best balance between stable trajectory exploitation and unstable regime suppression, while smaller values ($\lambda=5$) lead to over-smoothed results due to insufficient suppression of unstable predictions, and $\lambda=15$ introduces subtle discontinuity artifacts from overly sharp transitions. In our main experiments, we set $\lambda=10$ across all tasks.

\section{More Qualitative Results}
Rapid multimodal advances support diverse image applications, including hyperspectral classification~\cite{cui2022litedepthwisenet}, voice-supervised text spotting~\cite{tang2022voice}, and content-safety understanding~\cite{qiu2026yuvionvl}.

\begin{figure*}[t]
\centering
\includegraphics[width=1.0\textwidth]{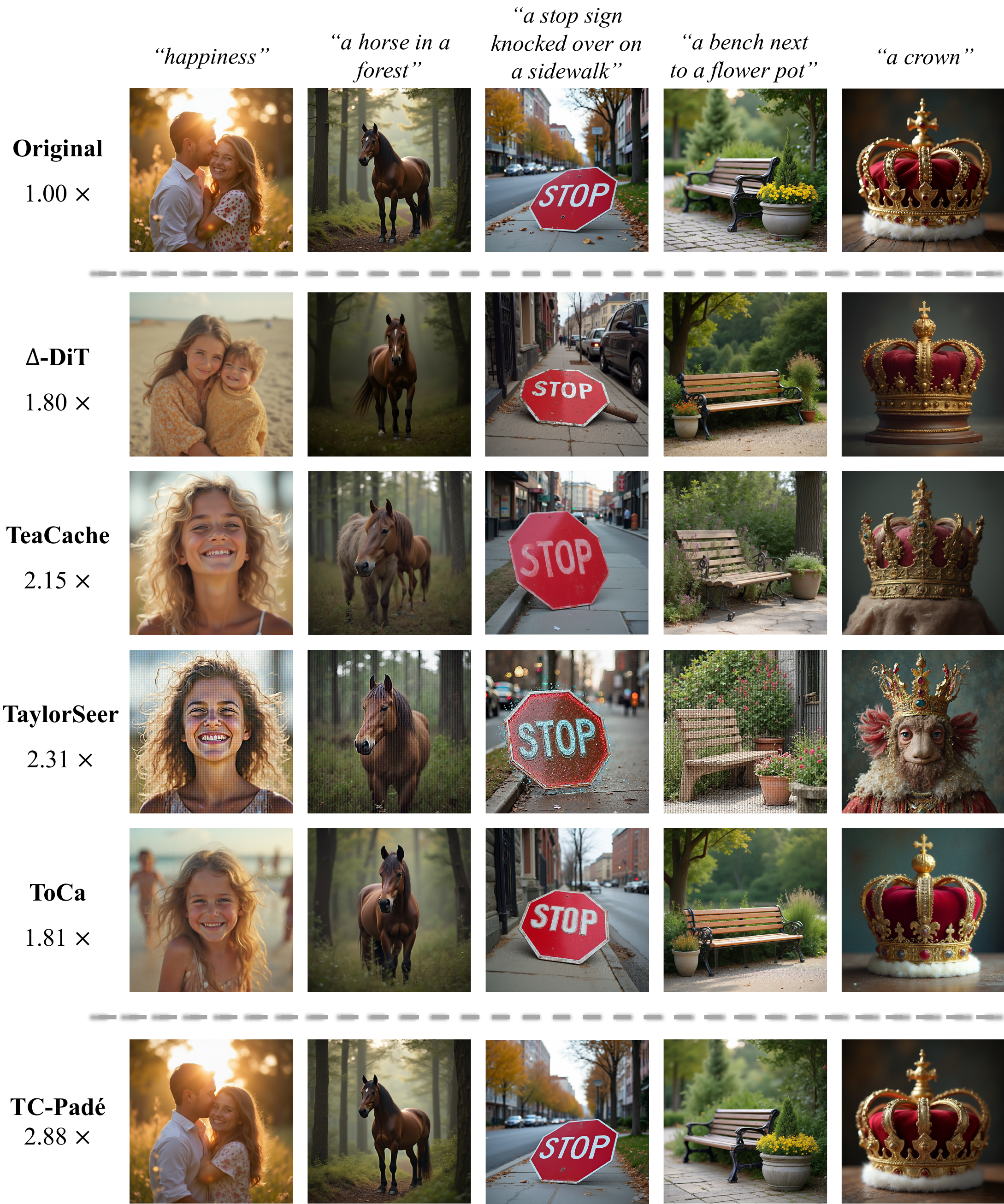}
\caption{
Qualitative examples for TC-Padé and other cache methods on FLUX.1-dev using 20 sampling steps. The text prompts are sampled from Parti prompts.
}
\label{fig:overview_flux_1}
\end{figure*}

\begin{figure*}[t]
\centering
\includegraphics[width=0.98\textwidth]{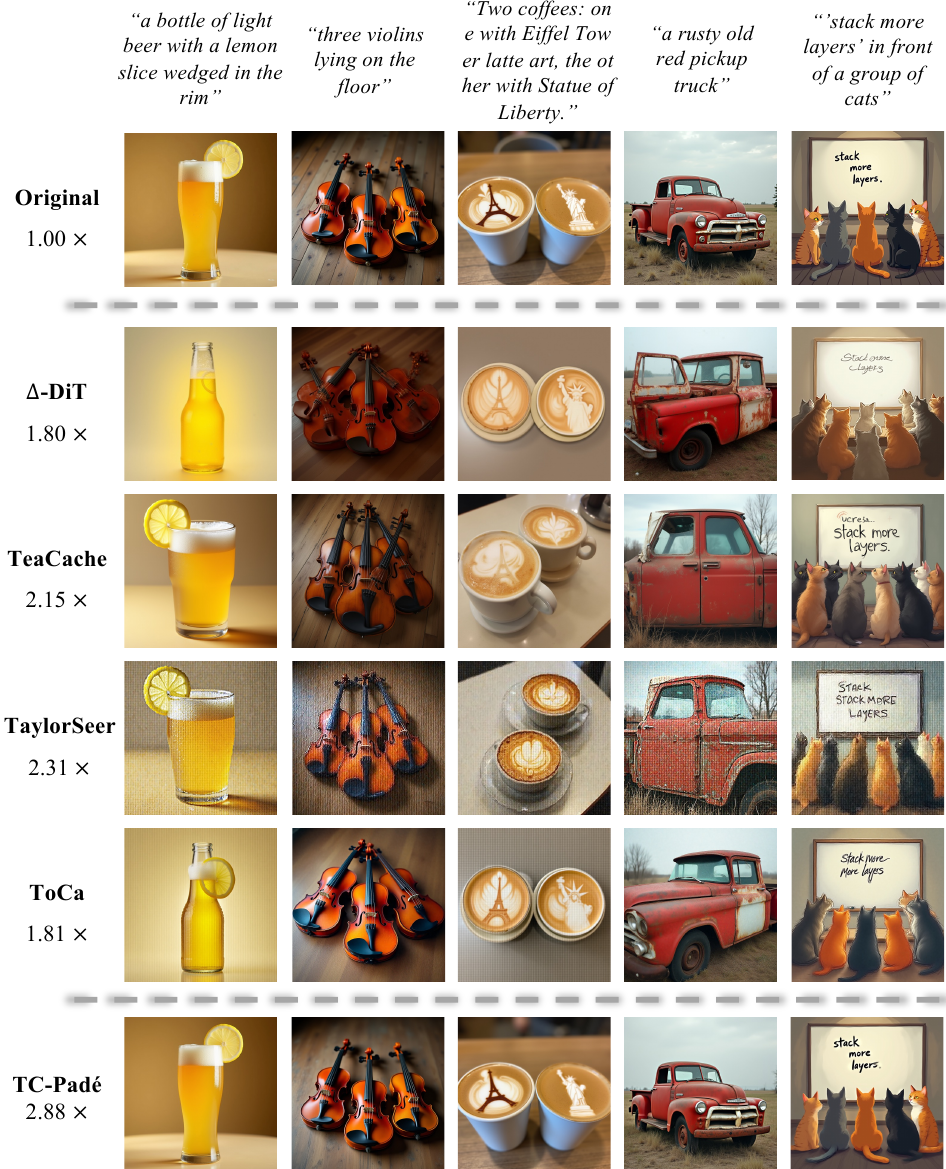}
\caption{
Qualitative examples for TC-Padé and other cache methods on FLUX.1-dev using 20 sampling steps. The text prompts are sampled from Parti prompts.
}
\label{fig:overview_flux_2}
\end{figure*}

\begin{figure*}[t]
\centering
\includegraphics[width=1.0\textwidth]{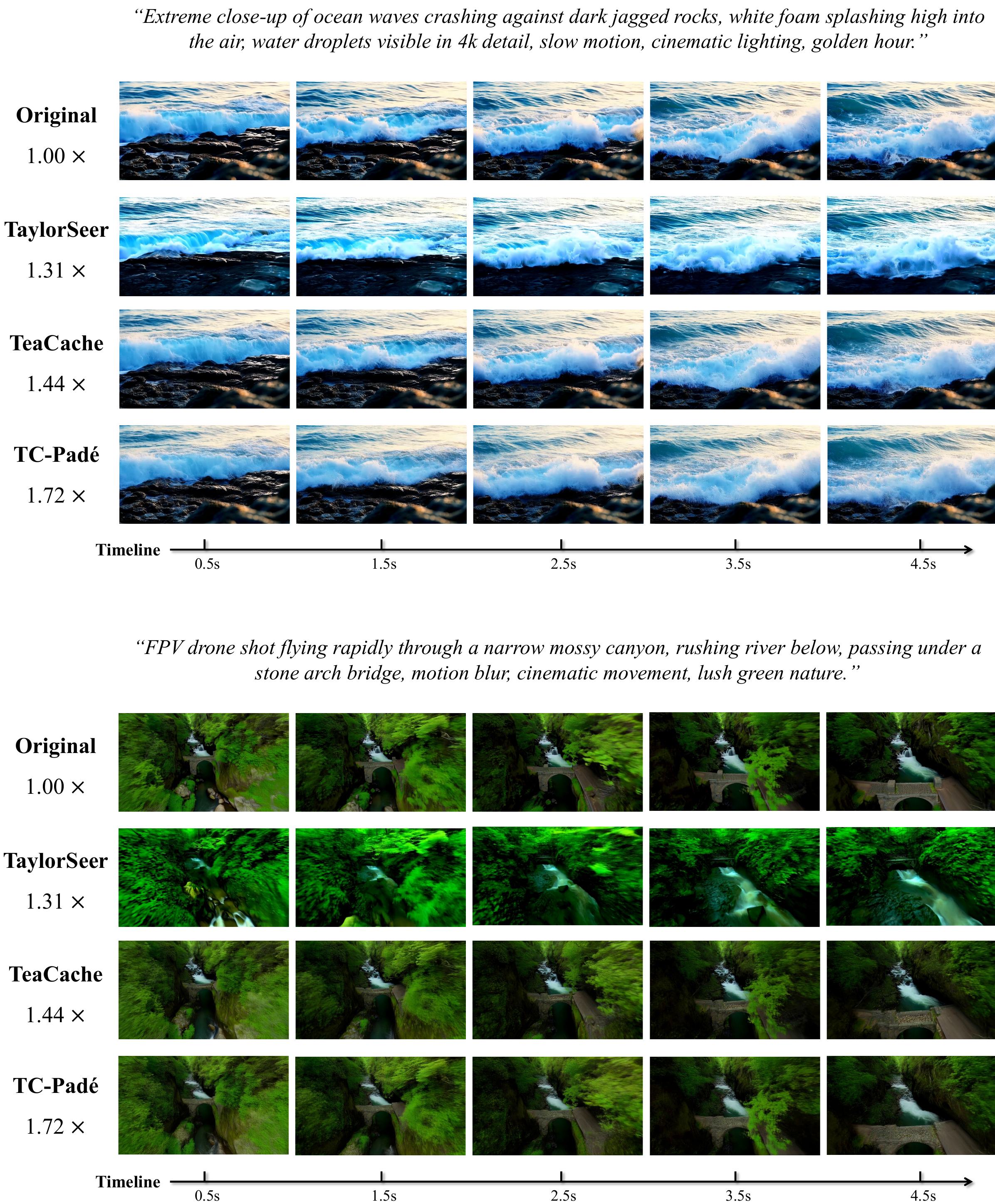}
\caption{
Qualitative examples for TC-Padé and other cache methods on Wan2.1-1.3B using 20 sampling steps.
}
\label{fig:overview_wan_1}
\end{figure*}

\begin{figure*}[t]
\centering
\includegraphics[width=1.0\textwidth]{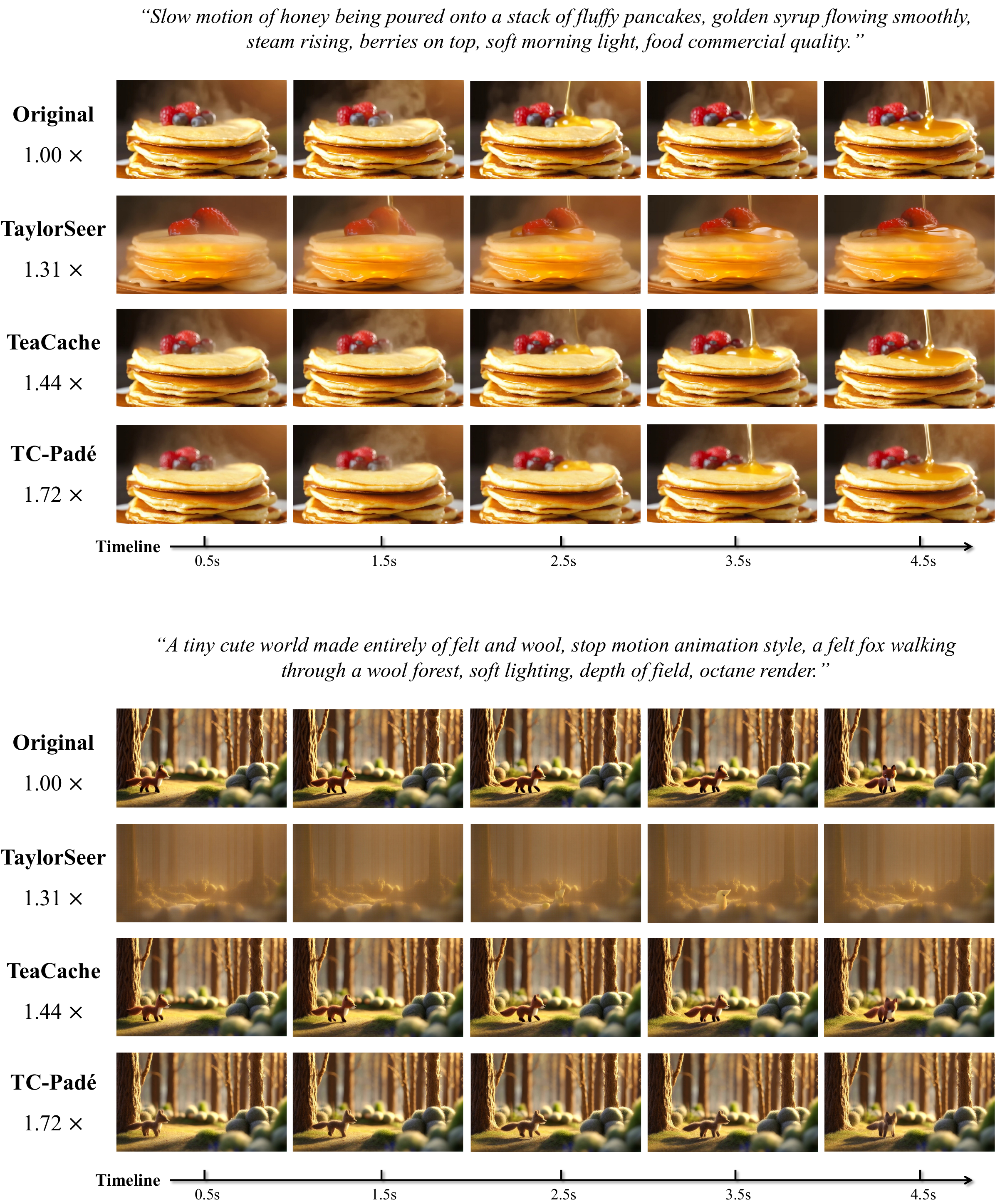}
\caption{
Qualitative examples for TC-Padé and other cache methods on Wan2.1-1.3B using 20 sampling steps.
}
\label{fig:overview_wan_2}
\end{figure*}

\end{document}